 \documentclass[accepted]{uai2023} 

\usepackage[american]{babel}

\usepackage{natbib} 
\usepackage{mathtools} 
\usepackage{booktabs} 
\usepackage{tikz} 

\DeclareMathOperator{\ReLU}{ReLU}
\DeclareMathOperator{\PReLU}{PReLU}
\newcommand{\ie}{\textit{i}.\textit{e}.}

\usepackage{cases}
\newtheorem{definition}{Definition}
\newtheorem{theorem}{Theorem}
\newtheorem{remark}{Remark}
\usepackage{derivative}
\usepackage{algorithm}
\usepackage{algpseudocode}
\usepackage{adjustbox}
\usepackage{subcaption}
\usepackage{graphicx}
\usepackage{booktabs}
\usepackage{multirow}

\title{On the Ideal Number of Groups for Isometric Gradient Propagation}

\author[1]{Bum Jun Kim}
\author[1]{Hyeyeon Choi}
\author[1]{Hyeonah Jang}
\author[1]{\href{mailto:<swkim@postech.edu>?Subject=Your UAI 2023 paper}{Sang Woo Kim}{}}
\affil[1]{%
    Department of Electrical Engineering.\\
    Pohang University of Science and Technology\\
    Pohang, South Korea
}

\begin{document}
\maketitle

\begin{abstract}
	Recently, various normalization layers have been proposed to stabilize the training of deep neural networks. Among them, group normalization is a generalization of layer normalization and instance normalization by allowing a degree of freedom in the number of groups it uses. However, to determine the optimal number of groups, trial-and-error-based hyperparameter tuning is required, and such experiments are time-consuming. In this study, we discuss a reasonable method for setting the number of groups. First, we find that the number of groups influences the gradient behavior of the group normalization layer. Based on this observation, we derive the ideal number of groups, which calibrates the gradient scale to facilitate gradient descent optimization. Our proposed number of groups is theoretically grounded, architecture-aware, and can provide a proper value in a layer-wise manner for all layers. The proposed method exhibited improved performance over existing methods in numerous neural network architectures, tasks, and datasets.
\end{abstract}

\section{Introduction}
\label{sec:introduction}
Deep neural networks have recently shown significant performance in various fields. Despite their current success, in the past, deep neural networks were known to be difficult to train. To stabilize the training of a deep neural network, normalization layers, such as batch normalization \citep{DBLP:conf/icml/IoffeS15}, have been proposed. Normalization layers have addressed the difficulty in the optimization of deep neural networks and are used in most deep neural networks at present.

Other widely used normalization layers include layer normalization \citep{DBLP:journals/corr/BaKH16}, instance normalization \citep{DBLP:journals/corr/UlyanovVL16}, and group normalization \citep{DBLP:journals/ijcv/WuH20}. These behave similarly in that they apply mean and standard deviation (std) normalization and an affine transform. The difference lies in the units used for computing the mean and std. For example, for $n$ features, layer normalization computes a single mean and std for normalization, whereas instance normalization computes $n$ means and stds. Meanwhile, group normalization partitions $n$ features into $G$ groups to compute $G$ means and stds. From this perspective, layer normalization is a special case of group normalization for $G = 1$, and instance normalization is a special case of group normalization for $G = n$. Thus, group normalization is more comprehensive and has a degree of freedom from the setting of the number of groups. When the number of groups is set to a specific value, there is a possibility of suboptimality, which leaves room for setting a more appropriate number of groups to further improve the performance.

The setting of the number of groups is also mentioned in the original paper on group normalization \citep{DBLP:journals/ijcv/WuH20}. By experimenting with several trials with $G = 1,\ 2,\ 4,\ \cdots$, they evaluated the ImageNet accuracy. They observed low accuracy at both extremes of $G=1$ and $G=n$. In particular, they empirically found the highest accuracy at $G=32$ and recommended this as the default value for the number of groups in group normalization. Accordingly, various studies using group normalization have employed $G=32$ \citep{DBLP:conf/cvpr/KirillovGHD19,DBLP:conf/nips/HoJA20,DBLP:conf/iclr/ZhuSLLWD21,DBLP:conf/iccv/YangLHWL19}.

However, this approach to setting the number of groups has several problems. First, the corresponding number of groups lacks theoretical validation. This is because the claim that $G=32$ yields the highest performance is confirmed only through empirical observations. Second, the neural network architecture is not considered. When a different architecture is employed, there is a possibility that $G=32$ is suboptimal; therefore, hyperparameter tuning by trial and error is required again to set the optimal number of groups. For example, \citet{DBLP:conf/cvpr/DaiCX0LY021} used $G=16$, whereas \citet{DBLP:journals/corr/abs-1908-00061} used $G=4$. Furthermore, \citet{DBLP:journals/tcsv/SongLK21} designed a neural network with a different number of groups for each layer. Training a deep neural network is time-consuming, and many hyperparameters already exist; employing an additional hyperparameter leads to a significantly high processing cost \citep{DBLP:conf/icml/FalknerKH18,DBLP:conf/nips/YangHBSLFRPCG21,DBLP:journals/prl/CuiB19}. Third, $G=32$ is not guaranteed to be optimal for all group normalization layers in a deep neural network using tens or hundreds of layers. In other words, since the optimal number of groups can be different for each layer, the number of groups in a layer-wise manner $G^l$ should be considered.

In this study, we propose an appropriate method for determining the number of groups. First, we theoretically analyze the effect of the number of groups on the back-propagation of group normalization. In this regard, we consider a gradient condition that facilitates the training of the neural network and derive the ideal number of groups that satisfies the gradient condition. Second, we show that the ideal number of groups we derived is affected by the width of the neural network. Hence, the ideal number of groups exhibits different values depending on the number of input and output features in the neural network architecture. Third, we demonstrate that the ideal number of groups varies for each layer. In summary, for setting the number of groups, we propose a reasonable method that is theoretically grounded, architecture-aware, and able to provide a proper value for all layers in a layer-wise manner.

For the application of the ideal number of groups, we propose the practical number of groups and apply it to several training experiments on deep neural networks. The proposed practical number of groups demonstrated higher performance in various tasks, architectures, and datasets.

\section{Ideal Number of Groups}
\label{sec:ontheidealnumberofgroups}
\subsection{Theoretical Analysis}
\label{sec:theoreticalanalysis}

\paragraph{Notation} In this paper, we use the notations $E[x_i]$ and $Var[x_i]$ to denote the mean and variance computed along the feature, $i$-axis. We do not use sample variance.\footnote{Some libraries apply Bessel's correction by default when measuring the variance. To obtain correct results such as those in Table \ref{tab:EqAD}, it should be turned off. For example, in PyTorch, \texttt{torch.var(input, unbiased=False)} should be used to apply a biased estimator. In fact, \texttt{unbiased=False} is specified when \texttt{torch.nn.GroupNorm()} measures the standard deviation.}

\begin{figure}[t!]
	\centering
	\includegraphics[width=0.99\columnwidth]{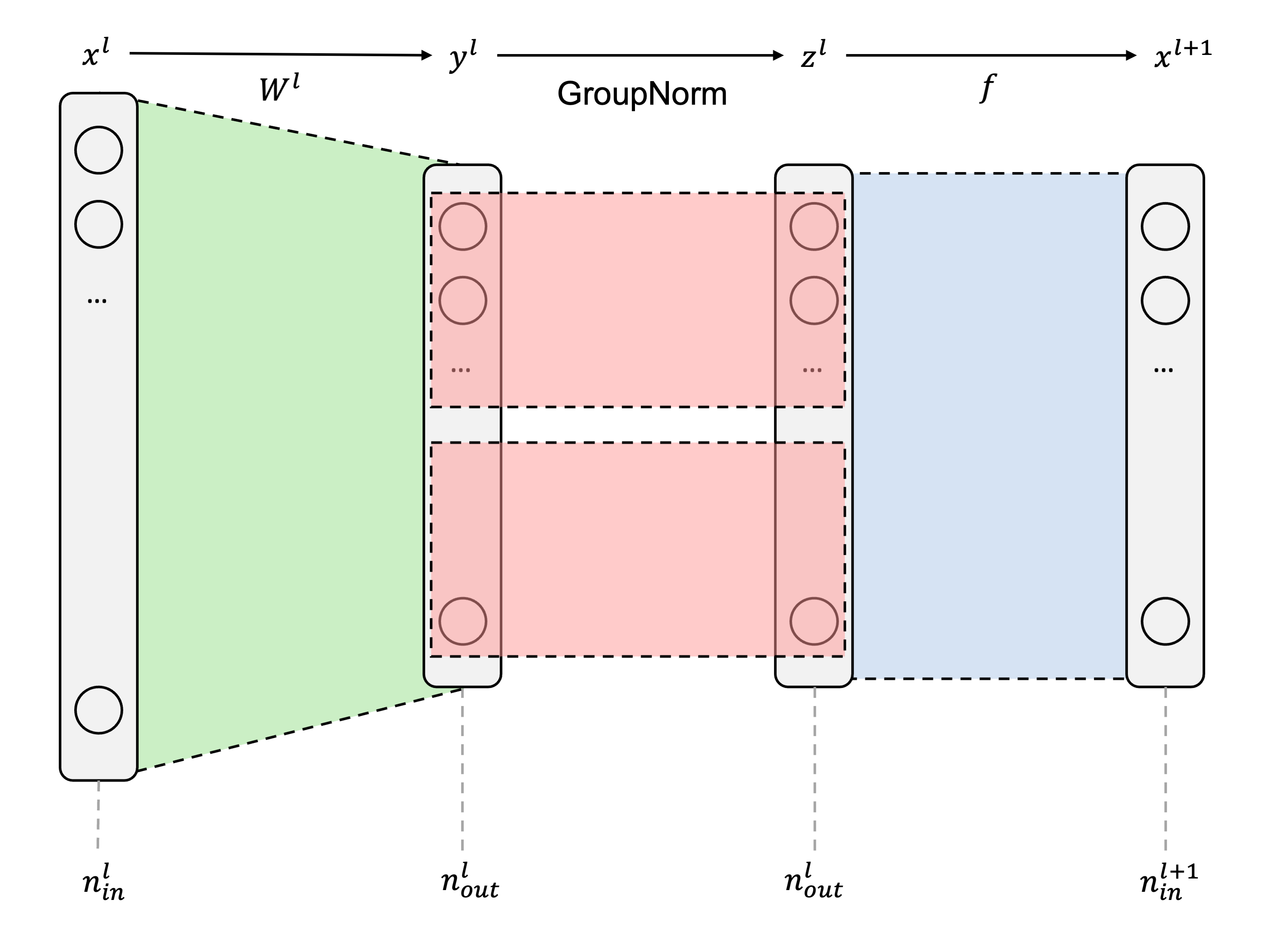}
	\caption{Illustration of the unit block.}
	\label{fig:unit}
\end{figure}

\paragraph{Formulation}
Consider a unit block that consists of a weight layer, group normalization, and ReLU activation function (Figure \ref{fig:unit}). First, we denote $n^l_{in}$-dimensional input features in the $l$-th block as $\mathbf{x}^l = \left(x^l_1,\ x^l_2,\ \cdots,\ x^l_{n^l_{in}}\right)$. Weight in the $l$-th block is denoted as $W^l \in R^{n^l_{in} \times n^l_{out}}$. We assume zero-mean weights, as \citet{DBLP:journals/jmlr/GlorotB10} and \citet{DBLP:conf/iccv/HeZRS15} did. The weight layer produces output feature $\mathbf{y}^l$, where
\begin{align}
	y^l_j = \sum_{i=1}^{n^l_{in}} W^l_{ij} x^l_i. \label{eq:weight}
\end{align}

Now, group normalization with the number of groups $G^l$ is applied to $\mathbf{y}^l$ to produce
\begin{align}
	z^l_{(s-1)n^l_g+j} = \frac{y^l_{(s-1)n^l_g+j} - \mu^l_s}{\sqrt{(\sigma^l_s)^2}}, \label{eq:gn}
\end{align}
where
\begin{align}
	\mu^l_s        & = \frac{1}{n^l_g} \sum_{j=1}^{n^l_g} y^l_{(s-1)n^l_g+j}, \label{eq:mu}                \\
	(\sigma^l_s)^2 & = \frac{1}{n^l_g} \sum_{j=1}^{n^l_g} (y^l_{(s-1)n^l_g+j}-\mu^l_s)^2, \label{eq:sigma} \\
	n^l_g          & = \frac{n^l_{out}}{G^l},
\end{align}
for $s=1,\ 2,\ \cdots,\ G^l$. The above equations mean partitioning $n^l_{out}$ features into $G^l$ groups and normalizing features in each group using the corresponding mean $\mu^l_s$ and std $\sigma^l_s$. In group normalization, an affine transform of $\gamma z^l_k+\beta$ is additionally used. Following \citet{DBLP:conf/nips/DeS20} and \citet{DBLP:conf/iclr/ZhangWXG19}, we use the default values, \ie, $\gamma=1$ and $\beta=0$. Finally, the activation function $f$ results in the next feature at the $(l+1)$-th block:
\begin{align}
	x^{l+1}_{i} = f(z^l_k). \label{eq:activation}
\end{align}
In summary, we obtain $\mathbf{x}^{l+1}$ from $\mathbf{x}^{l}$ by passing a unit block. Here, we consider the following property of the unit block.
\begin{definition}
	A unit block mapping $\mathbf{x}^l$ to $\mathbf{x}^{l+1}$ is \textit{isometric with respect to gradient propagation} if
	\begin{align}
		Var\Biggl[\frac{\partial L}{\partial x^{l+1}_i}\Biggr] = Var\Biggl[\frac{\partial L}{\partial x^l_i}\Biggr],
	\end{align}
	where $L$ denotes a loss function.
\end{definition}
This property ensures that the gradient scale is the same in both layers, which prevents unstable optimization due to exploding and vanishing gradients during gradient descent. For example, if the two variances are 10 and 1, this implies an imbalance in the gradient scale, which leads to an unstable optimization in the gradient descent. To stabilize the optimization, it is desirable to obtain the same or the most similar gradient scale. Note that the imbalance in the gradient scale is accumulated by passing tens or hundreds of unit blocks, which results in an exploding or vanishing gradient. So we aim to ensure that each unit block is isometric with respect to gradient propagation (Section \ref{sec:discussion}). This property was also the objective of \citet{DBLP:journals/jmlr/GlorotB10}, \citet{DBLP:conf/iccv/HeZRS15}, and \citet{DBLP:conf/nips/KlambauerUMH17}. In the remainder of our paper, unless specified otherwise, we use the term isometricity to discuss gradient propagation, not forward propagation.

Here, we claim that the number of groups affects the gradient variance. Our goal is to determine the solution for the number of groups that induces the isometric gradient propagation of the unit block. We investigate this, the ideal number of groups $G^l_{ideal}$.

\paragraph{Gradient propagation on weight layer}
First, from Eq. \ref{eq:weight}, note that an input feature $x^l_i$ affects all output features $y^l_1,\ y^l_2,\ \cdots,\ y^l_{n^l_{out}}$. From the chain rule for partial derivatives, we have
\begin{align}
	\frac{\partial L}{\partial x^l_i} = \sum_{j=1}^{n^l_{out}} \frac{\partial L}{\partial y^l_j} \frac{\partial y^l_j}{\partial x^l_i}.
\end{align}

By computing the variance, we see that $n_{out}^l$ components affect the variance:
\begin{align}
	Var\Biggl[\frac{\partial L}{\partial x^l_i}\Biggr] = n^l_{out} Var\left[W^l\right] Var\Biggl[\frac{\partial L}{\partial y^l_j}\Biggr]. \label{eq:dLdx}
\end{align}

\paragraph{Gradient propagation on group normalization}
Second, we investigate the backward propagation of group normalization. Notably, the gradients propagate only within the group. Consider the case in which a feature $y^l_j$ belongs to the $r$-th group. By Eqs. \ref{eq:mu} and \ref{eq:sigma}, we have
\begin{align}
	\frac{\partial \mu^l_r}{\partial y^l_j}          & = \frac{1}{n^l_g},                \\
	\frac{\partial [(\sigma^l_r)^2]}{\partial y^l_j} & = \frac{2(y^l_j-\mu^l_r)}{n^l_g}.
\end{align}

From Eq. \ref{eq:gn}, we find that the partial derivative differs depending on whether the index matches. We consider two cases:
\begin{numcases}{\frac{\partial z^l_k}{\partial{y^l_j}}=}
	\frac{1-\frac{1}{n^l_g}(1+(z^l_k)^2)}{\sigma^l_r},  & if $k = j$.  \\
	\frac{-\frac{1}{n^l_g}(1+z^l_k z^l_j)}{\sigma^l_r}, & if $k \neq j$.
\end{numcases}

From the chain rule for partial derivatives, we obtain
\begin{align}
	\frac{\partial L}{\partial y^l_j} & = \sum_{k=1}^{n^l_g} \frac{\partial L}{\partial z^l_k} \frac{\partial z^l_k}{\partial y^l_j} \\
	                                  & = T_1 - T_2 - T_3,
\end{align}
where
\begin{align}
	T_1 & = \frac{1}{\sigma^l_r}\left(\frac{\partial L}{\partial z^l_k}\right),                                                \\
	T_2 & = \frac{1}{\sigma^l_r}\frac{1}{n^l_g}\left(\sum_{k=1}^{n^l_g} \frac{\partial L}{\partial z^l_k}\right),              \\
	T_3 & = \frac{1}{\sigma^l_r}\frac{1}{n^l_g} z^l_j \left(\sum_{k=1}^{n^l_g} z^l_k \frac{\partial L}{\partial z^l_k}\right).
\end{align}

By computing the variance, we have
\begin{align}
	Var\left[T_1\right] & = \frac{1}{(\sigma^l_s)^2} Var\Biggl[\frac{\partial L}{\partial z^l_k}\Biggr], \label{eq:t1}                 \\
	Var\left[T_2\right] & = \frac{1}{(\sigma^l_s)^2} \frac{1}{n^l_g} Var\Biggl[\frac{\partial L}{\partial z^l_k}\Biggr], \label{eq:t2} \\
	Var\left[T_3\right] & = \frac{1}{(\sigma^l_s)^2} \frac{3}{n^l_g} Var\Biggl[\frac{\partial L}{\partial z^l_k}\Biggr]. \label{eq:t3}
\end{align}
The third equation holds because $E[(z^l_k)^2 \pdv{L}{z^l_k}]=0$ and $E[(z^l_k)^4] = 3$ for normalized feature $z^l_k$. We denote the $s$-th group to represent an arbitrary group. The variance is computed across all features, not the features within a group. Summarizing Eqs. \ref{eq:t1}-\ref{eq:t3}, we obtain
\begin{align}
	Var\Biggl[\frac{\partial L}{\partial y^l_j}\Biggr] = \frac{1}{(\sigma^l_s)^2} \Biggl(1+\frac{4}{n^l_g}\Biggr) Var\Biggl[\frac{\partial L}{\partial z^l_k}\Biggr]. \label{eq:dLdy}
\end{align}
Note that the number of groups $G^l$ is involved in here because $n^l_g = \frac{n^l_{out}}{G^l}$. Thus, the number of groups affects the gradient propagation on the group normalization layer. We exploit this fact as a key to configure the unit block to the state that is closest to isometric.

\paragraph{Gradient propagation on activation function}
Here, we investigate the activation function. To derive the variance around the activation function, we introduce the following two properties.
\begin{definition}
	Assume a random variable $X \sim \mathcal{N}(0,\ \sigma^2)$ and an arbitrary random variable $Y$. For a given activation function $f$, we define \textit{forward activation gain} $F_{f,\sigma}$ and \textit{backward activation gain} $B_{f,\sigma}$ as follows:
	\begin{align}
		F_{f,\sigma} & = \frac{E[(f(X))^2]}{Var[X]},         \\
		B_{f,\sigma} & = \frac{Var[f^{\prime}(X)Y]}{Var[Y]}.
	\end{align}
\end{definition}
In particular, if $E[Y]=0$, we have $B_{f,\sigma} = E[(f^{\prime}(X))^2]$.

\begin{remark} \label{rem:relu}
	If $f(x)=\ReLU(x)=\max(0,\ x)$, we have $F_{f,\sigma} = B_{f,\sigma} = \frac{1}{2}$.
\end{remark}
Especially for ReLU, the two gains are independent of $\sigma$. However, for the other activation functions, the two gains can vary depending on $\sigma$ (Section \ref{sec:otheractivationfunctions}).

Now we investigate the variance around the activation function. By Eq. \ref{eq:activation}, we see that
\begin{align}
	\pdv{L}{z^l_k} = \pdv{L}{x^{l+1}_i} \pdv{x^{l+1}_i}{z^l_k} = \pdv{L}{x^{l+1}_i} f^{\prime}(z^l_k).
\end{align}

Thus,
\begin{align}
	Var\Biggl[\pdv{L}{z^l_k}\Biggr] = B_{f,\sigma} Var\Biggl[\pdv{L}{x^{l+1}_i}\Biggr]. \label{eq:dLdz}
\end{align}

In addition, investigating the forward propagation of the $(l-1)$-th block,
\begin{align}
	(\sigma^l_s)^2 & = Var\left[y^l_j\right] = n^l_{in} Var\left[W^l f(z^{l-1}_i) \right] \\
	               & = n^l_{in} E\left[(W^l)^2\right] E\left[(f(z^{l-1}))^2\right]        \\
	               & = n^l_{in} F_{f,\sigma} Var\left[W^l\right].
\end{align}

\paragraph{Gradient propagation on unit block}
Finally, from Eqs. \ref{eq:dLdx}, \ref{eq:dLdy}, and \ref{eq:dLdz}, we obtain the gradient equation from $\mathbf{x}^l$ to $\mathbf{x}^{l+1}$:
\begin{align}
	Var\Biggl[\frac{\partial L}{\partial x^l_i}\Biggr] = \frac{n^l_{out}}{n^l_{in}} \Biggl(1+\frac{4}{n^l_g}\Biggr) Var\Biggl[\pdv{L}{x^{l+1}_i}\Biggr]. \label{eq:unitblockvar}
\end{align}

Let $K(G^l)$ be the ratio of two variances as
\begin{align}
	K(G^l) = \frac{n^l_{out}}{n^l_{in}} \Biggl(1+\frac{4}{n^l_g}\Biggr) = \frac{n^l_{out} + 4 G^l}{n^l_{in}}.
\end{align}
When $K(G^l) = 1$, the unit block is isometric with respect to gradient propagation. Thus, our goal is to find the number of groups $G^l$ that satisfies $K(G^l)=1$. Ideally, this condition can be satisfied if the group normalization has
\begin{align}
	G^l_{ideal} = \frac{n^l_{in} - n^l_{out}}{4},
\end{align}
which we call the ideal number of groups. Interestingly, the ideal number of groups depends on the architecture of the neural network, especially the number of input and output features of the weight layer. For example, for an $l$-th unit block with 128 input features and 64 output features, applying $G^l=16$ provides isometricity of the unit block. The use of this formula allows us to set an appropriate number of groups on a given layer or architecture without any tuning experiments. Intuitively, it is desirable to have a different number of groups depending on each width. In other words, it is unnatural to apply $G^l=32$ to all layers, regardless of the variety of width of the deep neural network.

\begin{algorithm}[t!]
	\caption{Compute $G^l_{practical}$} \label{alg:gpr}
	\begin{algorithmic}[1]
		\Require $n^l_{in}$ and $n^l_{out}$.
		\State Compute the ideal number of groups:

		$G^l_{ideal} = (n^l_{in}-n^l_{out})/4$.
		\State Apply lower bound: $G^l = \max(1, G^l_{ideal})$.
		\State Apply upper bound: $G^l = \min(G^l, n^l_{out})$.
		\State Compute the log of the number of groups: $\log_2 G^l$.
		\State Obtain divisor set of $n^l_{out}$ as $[n^l_{out}]$.
		\State Compute log of set $[n^l_{out}]$ as $\log_2[n^l_{out}]$.
		\State Find the closest value of $\log_2 G^l$ from the set $\log_2[n^l_{out}]$ as $l_g$.
		\State Compute the practical number of groups:

		$G^l_{practical} = 2^{l_g}$.
		\\
		\Return $G^l_{practical}$.
	\end{algorithmic}
\end{algorithm}

However, the ideal number of groups may not be applicable in a practical scenario depending on $n_{in}^l$ and $n_{out}^l$. For example, if $n_{in}^l=n_{out}^l=128$, then $G_{ideal}^l=0$, which cannot be employed in group normalization. Note that group normalization is applied to $n^l_{out}$ features. Thus, 1) the number of groups should have lower and upper bounds of $1 \leq G^l_{ideal} \leq n^l_{out}$, and 2) $n^l_{out}$ should be divisible by an integer $G^l_{ideal}$. Considering this, we seek a practical number of groups.

\begin{definition}
	Let $[n^l_{out}]$ be the divisor set of $n^l_{out}$. Find $G^l \in [n^l_{out}]$ where $K(G^l)$ is closest to 1. We refer to the result as the practical number of groups $G^l_{practical}$.
\end{definition}

\begin{table*}[t!]
	\caption{Empirical validation of Eqs. \ref{eq:A} to \ref{eq:D}. The results were in agreement with the theoretical expectations.}
	\label{tab:EqAD}
	\centering
	\resizebox{\linewidth}{!}{%
		\begin{tabular}{c|cc|cc|cc|cc}
			\toprule
			\multirow{2}{*}{($n^l_{in}$, $n^l_{out}$, $G^l$)} & \multicolumn{2}{c|}{Eq. \ref{eq:A}} & \multicolumn{2}{c|}{Eq. \ref{eq:B}} & \multicolumn{2}{c|}{Eq. \ref{eq:C}} & \multicolumn{2}{c}{Eq. \ref{eq:D}}                                                     \\
			                                                  & Empirical                           & Theoretical                         & Empirical                           & Theoretical                        & Empirical & Theoretical & Empirical & Theoretical \\
			\midrule
			(1024, 512, 128)                                  & 511.236                             & 512                                 & 1.979                               & 2                                  & 0.500     & 0.5         & 0.995     & 1           \\
			(512, 256, 64)                                    & 255.160                             & 256                                 & 1.959                               & 2                                  & 0.499     & 0.5         & 0.989     & 1           \\
			(256, 128, 32)                                    & 127.189                             & 128                                 & 2.015                               & 2                                  & 0.500     & 0.5         & 1.032     & 1           \\
			(128, 64, 16)                                     & 63.185                              & 64                                  & 1.900                               & 2                                  & 0.500     & 0.5         & 0.992     & 1           \\
			\bottomrule
		\end{tabular}
	}
\end{table*}

The objective of finding the practical number of groups is to configure the unit block to the state that is closest to isometric, facilitating gradient descent optimization. We consider three cases:

Case 1. If $n^l_{in} \leq n^l_{out}$, then $K(G^l)$ is always greater than one. Thus we seek $G^l$ that yields the smallest $K(G^l)$. Because $K(G^l)$ increases linearly with $G^l$, we choose the smallest $G^l$, \ie, $G^l_{practical}=1$. In fact, this is equivalent to applying a lower bound to the ideal number of groups.

Case 2. If $n^l_{in} \geq 5 n^l_{out}$, we have $K(G^l) = (n^l_{out} + 4 G^l) / n^l_{in} \leq 1$. We need to find $G^l$ that results in the highest $K(G^l)$. Thus, we choose $G^l_{practical}=n^l_{out}$. Similarly, this is equivalent to applying an upper bound to the ideal number of groups.

Case 3. If $n^l_{out} < n^l_{in} < 5n^l_{out}$, then we choose the number of groups in the divisor set $[n^l_{out}]$ that is closest to the ideal number of groups.

From the above analyses, we conclude with the following theorem.
\begin{theorem}
	Algorithm \ref{alg:gpr} yields the practical number of groups $G^l_{practical}$.
\end{theorem}

For example, if $n_{in}^l=n_{out}^l=128$, by Case 1, we choose $G^l_{practical}=1$. In this scenario, because $K(G^l) = 1+4G^l/128$, we obtain $K(1)=1.03125$. However, choosing a different number of groups such as 32 results in $K(32)=2$, which results in an imbalance in the gradient scale.

\subsection{Empirical Validation}
\label{sec:empiricalvalidation}
In this section, we test the validity of our derivation. We target Eqs. \ref{eq:dLdx}, \ref{eq:dLdy}, \ref{eq:dLdz}, and \ref{eq:unitblockvar}, which are our main results. The four equations are rewritten as follows:
\begin{align}
	Var\Biggl[\frac{\partial L}{\partial x^l_i}\Biggr] / Var\Biggl[\frac{\partial L}{\partial y^l_j}\Biggr]                & = n^l_{out} Var\left[W^l\right], \tag{A} \label{eq:A}                              \\
	(\sigma^l_s)^2 Var\Biggl[\frac{\partial L}{\partial y^l_j}\Biggr] / Var\Biggl[\frac{\partial L}{\partial z^l_k}\Biggr] & = 1+\frac{4}{n^l_g}, \tag{B} \label{eq:B}                                          \\
	Var\Biggl[\pdv{L}{z^l_k}\Biggr] / Var\Biggl[\pdv{L}{x^{l+1}_i}\Biggr]                                                  & = B_{f,\sigma}, \tag{C} \label{eq:C}                                               \\
	Var\Biggl[\frac{\partial L}{\partial x^l_i}\Biggr] / Var\Biggl[\pdv{L}{x^{l+1}_i}\Biggr]                               & = \frac{n^l_{out}}{n^l_{in}} \Biggl(1+\frac{4}{n^l_g}\Biggr). \tag{D} \label{eq:D}
\end{align}

First, we empirically measure the left-hand side of Eqs. \ref{eq:A} to \ref{eq:D} (Empirical) and compare the results with the right-hand side (Theoretical). We use two unit blocks, where $W^1 \in R^{n_{in} \times n_{in}}$. and $W^2 \in R^{n_{in} \times n_{out}}$ with unit variance. We generate artificial random data sampled from the standard normal distribution. The random data are provided to the first unit block, and we measure the ratios of variances targeting the second unit block. The loss function $L$ can be defined as an arbitrary function, and we simply define it as an aggregation of the output features of the second unit block. Four cases of ($n^l_{in}$, $n^l_{out}$, $G^l$) are tested. Considering randomness, for all results, we provide an average over $10^5$ experiments.

The results are summarized in Table \ref{tab:EqAD}. We observed that the empirical ratio of variances matched well with theoretical expectations.

\subsection{Other Activation Functions}
\label{sec:otheractivationfunctions}

\begin{table*}[th!]
	\caption{Empirical values for the forward and backward activation gains. For PReLU, slope $a=0.25$ was used. The six activation functions on the left consistently yielded $B_{f,\sigma} / F_{f,\sigma} \approx 1$, while the five activation functions on the right did not.
	}
	\label{tab:act}
	\centering
	\resizebox{\linewidth}{!}{%
		\begin{tabular}{l|cccccc|ccccc}
			\toprule
			Measurement             & ReLU  & PReLU & GELU  & SiLU  & ELU   & SELU  & Sigmoid & Tanh  & Softplus & Softsign & LogSigmoid \\
			\midrule
			$F_{f,0.1}$             & 0.500 & 0.531 & 0.255 & 0.252 & 0.928 & 1.876 & 25.079  & 0.981 & 48.438   & 0.751    & 48.495     \\
			$B_{f,0.1}$             & 0.500 & 0.531 & 0.256 & 0.253 & 0.929 & 1.879 & 0.062   & 0.981 & 0.251    & 0.757    & 0.251      \\
			$B_{f,0.1} / F_{f,0.1}$ & 1.000 & 1.000 & 1.006 & 1.002 & 1.001 & 1.001 & 0.002   & 1.000 & 0.005    & 1.007    & 0.005      \\
			\midrule
			$F_{f,1}$               & 0.500 & 0.532 & 0.425 & 0.356 & 0.645 & 1.000 & 0.293   & 0.394 & 0.921    & 0.183    & 0.921      \\
			$B_{f,1}$               & 0.500 & 0.532 & 0.456 & 0.379 & 0.668 & 1.071 & 0.045   & 0.464 & 0.293    & 0.228    & 0.293      \\
			$B_{f,1} / F_{f,1}$     & 1.000 & 1.000 & 1.072 & 1.067 & 1.036 & 1.072 & 0.153   & 1.178 & 0.318    & 1.245    & 0.319      \\
			\midrule
			$F_{f,10}$              & 0.500 & 0.531 & 0.500 & 0.499 & 0.504 & 0.565 & 0.005   & 0.009 & 0.501    & 0.007    & 0.501      \\
			$B_{f,10}$              & 0.500 & 0.531 & 0.506 & 0.507 & 0.520 & 0.613 & 0.007   & 0.053 & 0.461    & 0.026    & 0.461      \\
			$B_{f,10} / F_{f,10} $  & 1.000 & 1.000 & 1.011 & 1.017 & 1.032 & 1.085 & 1.434   & 5.780 & 0.920    & 3.908    & 0.921      \\
			\bottomrule
		\end{tabular}
	}
\end{table*}

\begin{table*}[t!]
	\caption{Test error (\%) for MNIST classification. Lower is better. $^*$ indicates the practical number of groups.}
	\label{tab:mnist}
	\centering
	\begin{tabular}{l|cccccccccc}
		\toprule
		$G^1$ & 1     & 2     & 4     & 8     & 16    & 32    & 64$^*$         & 128   & 256   & 512    \\
		\midrule
		Error & 1.827 & 1.720 & 1.773 & 1.730 & 1.727 & 1.720 & \textbf{1.670} & 1.753 & 3.240 & 88.650 \\
		\bottomrule
	\end{tabular}
\end{table*}

In this section, we investigate whether the ideal number of groups is applicable for other activation functions. First, consider $\PReLU(x) = \max(0,\ x) + a \min(0,\ x)$ with scalar $a$, which is a generalization of ReLU and LeakyReLU \citep{DBLP:conf/iccv/HeZRS15,maas2013rectifier}. We know that
\begin{align}
	E[(f(X))^2] & = \int_{-\infty}^{\infty}{(f(x))^2 p(x) dx}                           \\
	            & = \int_{-\infty}^{0}{(ax)^2 p(x) dx} + \int_{0}^{\infty}{x^2 p(x) dx} \\
	            & = \frac{1+a^2}{2} E[X^2],
\end{align}
where $p(x)$ denotes the probability density function of $X$. Similarly, $E[(f^{\prime}(X))^2] = \frac{1+a^2}{2}$. Thus, PReLU has $F_{f,\sigma} = B_{f,\sigma} = \frac{1+a^2}{2}$. Remark \ref{rem:relu} in Section \ref{sec:theoreticalanalysis} can be explained by $a=0$. Moreover, the PReLU family guarantees consistent gain.

\begin{remark} \label{rem:homo}
	The forward and backward activation gains do not vary by $\sigma$ if and only if the activation function $f$ is homogeneous $f(kx)=kf(x)$, or $f(kx)=-kf(x)$ for a scalar $k \neq 0$.
\end{remark}
See the Appendix for a detailed proof. For example, $\ReLU(x)$ and its mirror $-\ReLU(x)$ share two gains that are independent of $\sigma$.

Remark \ref{rem:homo} implies that the forward and backward gains vary by $\sigma$ for other activation functions, such as SiLU and ELU \citep{DBLP:conf/iclr/RamachandranZL18,DBLP:journals/nn/ElfwingUD18,DBLP:journals/corr/ClevertUH15}. Furthermore, their nonlinear exponential terms make it difficult to compute the exact solution of the forward and backward activation gains.

Alternatively, we provide empirical values for these two gains. We generate $10^7$ samples of $X \sim \mathcal{N}(0,\ \sigma^2)$ and measure the forward and backward activation gains on well-known activation functions \citep{hendrycks2016gaussian,DBLP:conf/nips/KlambauerUMH17,DBLP:conf/ijcnn/ZhengYLLL15,elliott1993better}. Here, we list the results for $\sigma$ of $\{0.1,\ 1,\ 10\}$ with various activation functions (Table \ref{tab:act}).

Note that $\frac{B_{f,\sigma}}{F_{f,\sigma}}=1$ is used in Eq. \ref{eq:unitblockvar} assuming ReLU. Some activation functions, such as ReLU, PReLU, GELU, SiLU, ELU, and SELU, yielded a value near 1; thus, the practical number of groups can be safely used with these activation functions. However, $\frac{B_{f,\sigma}}{F_{f,\sigma}}$ from other activation functions, such as Sigmoid, Tanh, Softplus, Softsign, and LogSigmoid, were far from 1. If we consider this, for an arbitrary activation function, the ideal number of groups should be $G^l_{ideal} = (\frac{F_{f,1}}{B_{f,1}}n^l_{in} - n^l_{out})/4$. See the Appendix for more results and discussions.

\section{Experiments}
\label{sec:experiments}

\subsection{Image Classification with MLP}

\begin{figure}[t!]
	\centering
	\includegraphics[width=0.99\columnwidth]{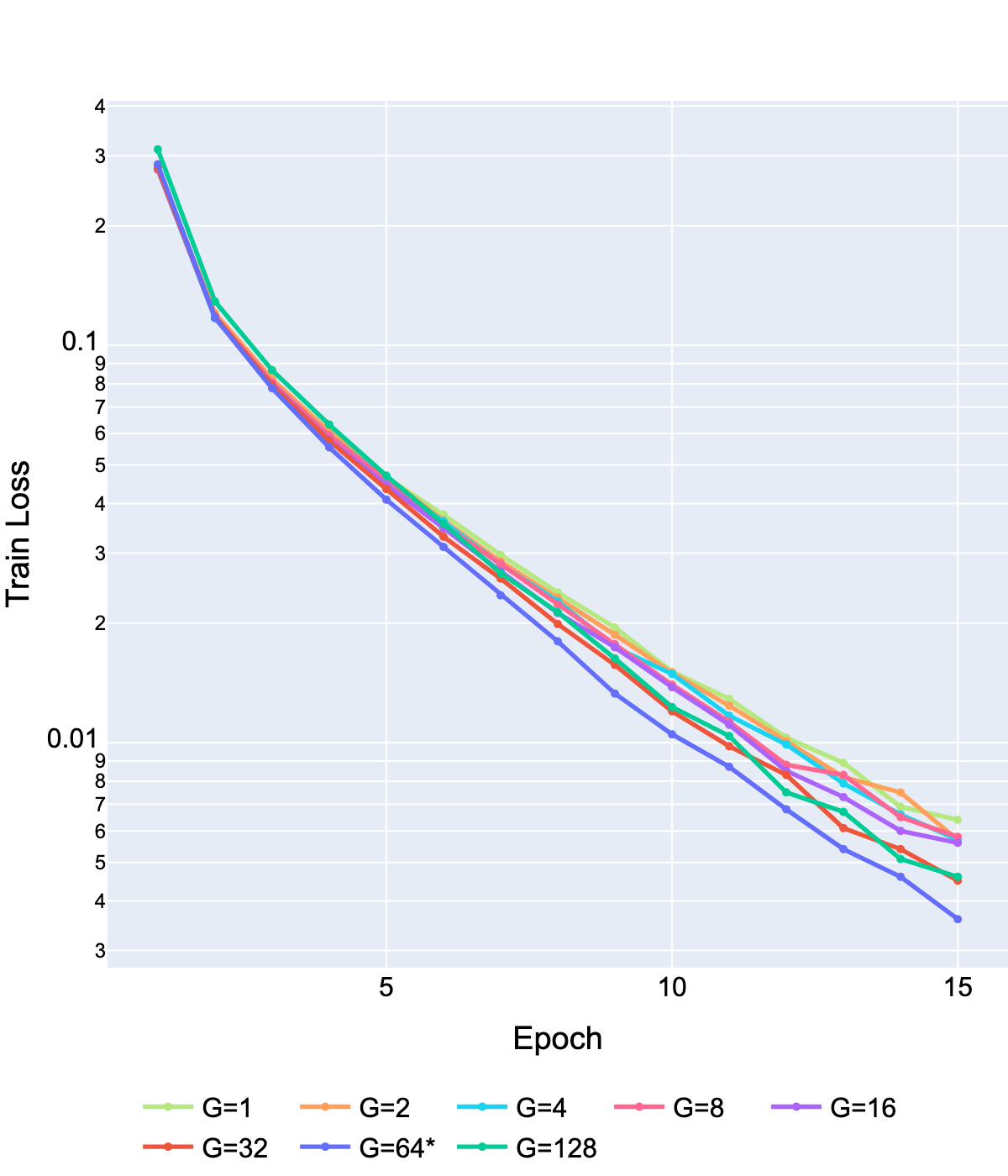}
	\caption{The learning curve of MLP for each number of groups.}
	\label{fig:traj}
\end{figure}

In this section, we aim to observe the performance differences of neural networks with different settings for the number of groups. We start with a simple task and then proceed to other large-scale tasks. First, we trained multilayer perceptrons (MLPs) for MNIST image classification. The simplicity of this task allows us to experiment with extensive numbers of groups. We used a two-layer MLP of 512 hidden nodes with group normalization and ReLU. Because $n_{in}^1=784$ for MNIST data and $n_{out}^1=512$, we have $G_{ideal}^1=68$ and thus $G_{practical}^1=64$.

An average over three runs is reported for each result (Table \ref{tab:mnist}). The highest accuracy was found in $G^1=64$, which corresponds to $G_{practical}^1$. Others, such as $G^1=32$ and $G^1=1$, worked fine but they left with possible improvements. Figure \ref{fig:traj} shows the learning curve of MLP for each number of groups. Because other conditions are the same, we can see that choosing the number of groups affected the learning curve of train loss, where $G^l=64$ yielded faster convergence compared to others. We conclude that the improved accuracy came from the improved optimization achieved by choosing $G^l=64$.

\subsection{Image Classification with ResNet}
\label{sec:imageclassificationwithresnet}
Second, we conducted experiments on convolutional neural networks (CNNs) for image classification. We targeted ResNet \citep{DBLP:conf/cvpr/HeZRS16}, which is a standard model used for image classification. After replacing the existing batch normalization layers with group normalization, we compared the performance with different numbers of groups. Because ResNet uses tens or hundreds of layers and is computationally expensive to train, we compared performance from three settings of the number of groups: $G^l=32$ as the default value of group normalization, $G^l=1$ corresponding to layer normalization, and $G^l=G^l_{practical}$ we proposed.

We targeted two datasets: Oxford-IIIT Pet and Caltech-101 \citep{DBLP:conf/cvpr/ParkhiVZJ12,DBLP:journals/cviu/Fei-FeiFP07}. The Oxford-IIIT Pet dataset includes 7K pet images of 37 classes, and the Caltech-101 dataset includes 9K object images of 101 classes with a background category. See the Appendix for details on experiments, such as the used hyperparameters. An average over three runs is reported for each result (Table \ref{tab:resnet}).

\begin{table}[]
	\caption{Test error (\%) for image classification. ``R'' represents ResNet. Lower is better.}
	\label{tab:resnet}
	\centering
	\begin{tabular}{l|c|c|c|c}
		\toprule
		\multirow{2}{*}{Setup} & \multicolumn{2}{c|}{Oxford-IIIT Pet} & \multicolumn{2}{c}{Caltech-101}                                     \\
		                       & R-50                                 & R-101                           & R-50            & R-101           \\
		\midrule
		$G^l=32$               & 22.894                               & 24.067                          & 24.021          & 22.781          \\
		$G^l=1$                & 33.514                               & 34.567                          & 24.167          & 26.647          \\
		$G^l=G^l_{practical}$  & \textbf{21.119}                      & \textbf{22.924}                 & \textbf{22.368} & \textbf{22.247} \\
		\bottomrule
	\end{tabular}
\end{table}

We observed that when $G^l=G^l_{practical}$ was employed, it achieved a higher accuracy than when $G^l=32$ or $G^l=1$. The performance improvement was consistently confirmed in the two datasets and ResNet-$\{50,\ 101\}$.

\subsection{Panoptic Segmentation with PFPN}
\label{sec:panopticsegmentationwithpfpn}
Panoptic segmentation is a task that simultaneously solves semantic and instance segmentation \citep{DBLP:conf/cvpr/KirillovHGRD19}. In other words, panoptic segmentation performs both pixel-wise classification and instance delineation. It is a large-scale downstream task that uses a CNN. Here, we focus on the panoptic feature pyramid network (PFPN), one of the representative models employed in the panoptic segmentation task \citep{DBLP:conf/cvpr/KirillovGHD19}. In addition, the PFPN originally exploited group normalization with $G^l=32$. We compare the performance of the PFPN for $G^l=32$, $G^l=1$, and $G^l=G^l_{practical}$.

The COCO-panoptic dataset \citep{DBLP:conf/eccv/LinMBHPRDZ14}, which includes labeled 80 things and 53 stuff, was used for training and testing. We measured the panoptic quality ($\mathrm{PQ}$), a commonly used performance index for the task \citep{DBLP:conf/cvpr/KirillovHGRD19}, and its variants $\mathrm{PQ^{th}}$ and $\mathrm{PQ^{st}}$ for thing and stuff, respectively (Table \ref{tab:pfpn}). The use of $G^l=G^l_{practical}$ resulted in a higher $\mathrm{PQ}$ than with $G^l=32$ or $G^l=1$. In particular, in the three indices, neither $G^l=32$ nor $G^l=1$ showed a clearly superior result, but for $G^l=G^l_{practical}$, higher $\mathrm{PQ}$ was consistently observed in all three indices.

\begin{table}[]
	\caption{Panoptic segmentation results, where a higher number is better.}
	\label{tab:pfpn}
	\centering
	\begin{tabular}{l|c|c|c}
		\toprule
		Setup                 & $\mathrm{PQ}$   & $\mathrm{PQ^{th}}$ & $\mathrm{PQ^{st}}$ \\
		\midrule
		$G^l=32$              & 41.750          & 49.357             & 30.268             \\
		$G^l=1$               & 41.461          & 49.688             & 29.043             \\
		$G^l=G^l_{practical}$ & \textbf{42.147} & \textbf{49.816}    & \textbf{30.572}    \\
		\bottomrule
	\end{tabular}
\end{table}

\subsection{Object Detection with Faster R-CNN GN+WS}
\label{sec:objectgdetectionwithfasterrcnn}

\begin{table}[]
	\caption{Object detection results for Faster R-CNN GN+WS, where a higher number is better.}
	\label{tab:faster}
	\centering
	\begin{tabular}{l|c|c|c}
		\toprule
		Setup                 & $\mathrm{AP}$ & $\mathrm{AP^{50}}$ & $\mathrm{AP^{75}}$ \\
		\midrule
		$G^l=32$              & 40.5          & 61.0               & 44.2               \\
		$G^l=1$               & 40.4          & 60.9               & 44.3               \\
		$G^l=G^l_{practical}$ & \textbf{40.7} & \textbf{61.2}      & \textbf{44.6}      \\
		\bottomrule
	\end{tabular}
\end{table}

\citet{qiao2019micro} suggested that when group normalization is used, improved training is possible when weight standardization is applied. They experimented with a combination of group normalization and weight standardization for various tasks. Motivated by this practice, we tested the application of the practical number of groups for the case of using group normalization and weight standardization.

The target model was Faster R-CNN with group normalization and weight standardization (GN+WS), which is an improved variant of Faster R-CNN, where the existing batch normalization layers are replaced with group normalization, and weight standardization is applied in the convolution layers \citep{DBLP:journals/pami/RenHG017,qiao2019micro}. The target task was object detection, which is a representative downstream task using a CNN. For training and testing, we used the COCO 2017 dataset, which consists of 118K training images, 5K validation images, and 41K test images. Average precision ($\mathrm{AP}$), which is a commonly used index, and its variants ($\mathrm{AP^{50}}$ and $\mathrm{AP^{75}}$) at $\mathrm{IoU}=50$ and $\mathrm{IoU}=75$ were measured (Table \ref{tab:faster}). Applying $G^l=G^l_{practical}$ resulted in minor but consistent improvements compared to $G^l=32$ or $G^l=1$.

\section{Discussion}
\label{sec:discussion}
\citet{DBLP:journals/jmlr/GlorotB10} discussed the condition under which the variance becomes equal for forward and backward propagation. Assuming a neural network composed of weight and sigmoid layers, for forward and backward propagation, they derived the following two conditions:
\begin{align}
	n^l_{in} Var\left[W^l\right] = 1,\ n^l_{out} Var\left[W^l\right] = 1. \label{eq:xavier_backward}
\end{align}

However, when $n^l_{in} \neq n^l_{out}$, because both conditions cannot be simultaneously satisfied, they proposed $Var\left[W^l\right] = \frac{2}{n^l_{in} + n^l_{out}}$ as a compromise to get as close to the two conditions as possible. This is applied at the initialization of the neural network to control the weights to attain the corresponding variance. \citet{DBLP:conf/iccv/HeZRS15} proposed another initialization method that considers the use of ReLU. These studies provide several notable points.

\paragraph{Perfect isometricity is not required.} As mentioned above, it is difficult to equalize the variance in the forward and backward propagation simultaneously. In other words, it is difficult to obtain isometricity with respect to both forward and backward propagations. \citet{DBLP:journals/jmlr/GlorotB10} presented a compromising alternative, and \citet{DBLP:conf/iccv/HeZRS15} considered only forward variance. Furthermore, even if the neural network was isometric at initialization, the isometricity would vanish during training. For example, weight decay reduces the weight norm during training, which causes it to lose isometricity. In summary, the initialization method of \citet{DBLP:journals/jmlr/GlorotB10} and \citet{DBLP:conf/iccv/HeZRS15} helps in training by making the neural network partially isometric but does not pursue perfect isometricity.

\paragraph{Their assumptions differ from the architectures of practical neural networks.} \citet{DBLP:journals/jmlr/GlorotB10} assumed that a neural network comprises a combination of weight layers and sigmoid activation functions. For this scenario, they derived the consecutive accumulation of backward variance from the $l$-th to $l^{\prime}$-th layer as
\begin{align}
	Var\Biggl[\frac{\partial L}{\partial x^l_i}\Biggr] = Var\Biggl[\frac{\partial L}{\partial x^{l^{\prime}}_i}\Biggr] \prod_{m=l}^{l^{\prime}-1} n^m_{out} Var\left[W^m\right]. \label{eq:xavier_scenario}
\end{align}

Thus, if Eq. \ref{eq:xavier_backward} holds for each layer, then Eq. \ref{eq:xavier_scenario} is satisfied, which makes the entire neural network isometric. Similarly, in our paper, we discussed the isometricity of a unit block composed of a weight layer, group normalization, and ReLU activation function. If a neural network is composed of only these unit blocks without any other operations, the use of an ideal number of groups will ensure isometricity for the entire neural network. However, \citet{DBLP:journals/corr/MishkinM15} argued that other operations, such as the maxpool operation and other activation functions, should be considered in practice. Because it is difficult to deal with all cases theoretically, they proposed normalizing the variance after empirically measuring it in a neural network. Practical neural networks include various operations such as strided operations and skip or dense connections. When these operations are exploited in conjunction with unit blocks in a neural network, it is difficult to conclude that the isometricity of the unit block guarantees the isometricity of the entire neural network.

Despite these limitations, the initialization methods of \citet{DBLP:journals/jmlr/GlorotB10} and \citet{DBLP:conf/iccv/HeZRS15} have been successfully deployed in modern neural networks. The initialization method of \citet{DBLP:conf/iccv/HeZRS15} is always specified in various libraries, including \texttt{torchvision.models}, pytorch image models (\texttt{timm}), and \texttt{MMClassification}. Several studies have reported the effectiveness of initialization methods for stable training \citep{DBLP:conf/iccv/HeGD19,DBLP:conf/aaai/ShangCS17,DBLP:conf/cvpr/HeZ0ZXL19}. These practices imply advocacy of partial isometricity for performance gain rather than opposition due to the limitation of partial isometricity.

In summary, these two viewpoints indicate that training is stabilized even if the neural network is 1) close to an isometric state rather than in a perfect isometric state and 2) isometric only in the local unit block. Similarly, our practical number of groups 1) is different from the ideal number of groups, so it is not a perfect solution and only partially makes it as isometric as possible and 2) guarantees isometricity only for the unit block, not the entire neural network. In other words, our practical number of groups provides only partial isometricity, but it is sufficient to facilitate training.

\section{Conclusion}
\label{sec:conclusion}
In this study, we proposed a practical method for determining the number of groups for group normalization. We stated the limitations of the trial-and-error-based hyperparameter tuning approach for setting the number of groups in group normalization. In this regard, we derived the ideal number of groups, which is advantageous for gradient descent optimization. Then we proposed the practical number of groups and applied it to various tasks, including image classification, panoptic segmentation, and object detection. We confirmed that the use of the practical number of groups provides improved performance compared to using the other settings for the number of groups.

\bibliography{bib}
\end{document}